\documentclass{article}

\PassOptionsToPackage{numbers}{natbib}
\usepackage[final]{nips_2018}

\usepackage[utf8]{inputenc} 
\usepackage[T1]{fontenc}    
\usepackage{hyperref}       
\usepackage{url}            
\usepackage{booktabs}       
\usepackage{amsfonts}       
\usepackage{nicefrac}       
\usepackage{microtype}      
\usepackage{graphicx}
\usepackage{authblk}

\title{Deep Meditations:\\Controlled navigation of latent space}

\author[1]{Memo Akten (\texttt{m.akten@gold.ac.uk})}
\author[1]{Rebecca Fiebrink (\texttt{r.fiebrink@gold.ac.uk})}
\author[2]{Mick Grierson (\texttt{m.grierson@arts.ac.uk})}
\affil[1]{Department of Computing, Goldsmiths University of London}
\affil[2]{Creative Computing Institute, University of the Arts, London}

\begin{document}

\maketitle

\begin{abstract}
We introduce a method which allows users to creatively explore and navigate the vast latent spaces of deep generative models. Specifically, our method enables users to \textit{discover} and \textit{design} \textit{trajectories} in these high dimensional spaces, to construct stories, and produce time-based media such as videos---\textit{with meaningful control over narrative}. Our goal is to encourage and aid the use of deep generative models as a medium for creative expression and story telling with meaningful human control. Our method is analogous to traditional video production pipelines in that we use a conventional non-linear video editor with proxy clips, and conform with arrays of latent space vectors. Examples can be seen at \url{http://deepmeditations.ai}.
\end{abstract}

\section{Introduction}
In recent years we have seen major progress in the capability of generative deep neural networks that are able to train on vast datasets and produce high resolution images. The goal of these generative models is to \textit{learn the distribution} of the training data such that we can sample from the distribution to generate new images. We refer to this distribution as the \textit{latent space} of the model, where any point in this space can be decoded to a unique image. We denote vectors in this space with \textbf{z}.

Depending on our goals, there are different ways in which we could explore such a space. We could use a fully automated search whereby points of interest are found algorithmically based on heuristics, such as novelty \cite{Lehman2011}. Alternatively, we could provide a target image and retrieve a corresponding \textbf{z}. Some architectures---e.g. Variational Auto-Encoders (VAE) \cite{Kingma2013} or Glow \cite{Kingma2018}---have an \textit{encoder} which does this, while other architectures---e.g. Generative Adversarial Networks (GAN) \cite{Goodfellow2014a}---do not. However it is still possible either by manually adding an encoder---e.g. VAE/GAN \cite{Dk2014}, or using gradient based optimization techniques to recover a corresponding vector \cite{Lipton2017b}. Other methods for exploring latent spaces include systematic visualizations of interpolations between key points \cite{White2016}.

Our paper is not an attempt to replace any of these methods for discovering interesting \textit{points} in latent space. Rather, we can incorporate them into our process to enable users to \textit{discover} and \textit{design} interesting \textit{trajectories} in latent space, to produce sequences with meaningful control over narrative. 

We face a number of challenges:
i) Latent spaces are vast and high dimensional (e.g. 512); ii) They are not distributed `evenly' or as one might expect or desire. If we were to sample uniformly across the space, we might end up with many more images of one type over another (e.g. in our model, flowers occupy a large portion of the space); iii) As a result, interpolating from $\textbf{z}_A$ to $\textbf{z}_B$ at a \textit{constant} speed might result in \textit{visually variable} speeds in movement; iv) It is very difficult to anticipate trajectories in high dimensions. E.g. interpolating from $\textbf{z}_A$ to $\textbf{z}_B$ might pass through $\textbf{z}_X$ and $\textbf{z}_Y$, which may be undesirable; v) the mass of the distribution is concentrated in the shell of a hypersphere; and vi) The latent space changes with subsequent training iterations. We discuss these in more detail below. 

\section{Method}
\paragraph{Overview:}  

It is common practice in video editing and post-production pipelines to perform an \textit{offline} edit on a set of proxy clips (e.g. low quality video), and later \textit{conform} the edit by applying it to \textit{online} material (e.g. high quality video). The conforming process is usually performed by transferring the edit to an \textit{online} system using a file such as an \textit{Edit Decision List (EDL)} which contains reel and timecode information as to where each video clip can be found in order to perform the final cut.

Our method is similar. Our \textit{offline} clips are the \textit{video} outputs from a generative model for given \textbf{z}-sequences. Our \textit{online} clips are numpy arrays of corresponding \textbf{z}-sequences. In other words, we define \textit{(\textbf{z}-sequence, video)} pairs. A \textit{\textbf{z}-sequence} is a numpy array (saved to disk) containing a sequence of \textbf{z}-vectors, i.e. \textit{a trajectory in latent space}. A \textit{video} is a QuickTime file where each frame is the image output from the model decoding the corresponding \textbf{z} from the corresponding \textbf{z}-sequence. We edit videos in a Non Linear Video Editor (NLE), and then run a custom script to conform the edit with the corresponding \textbf{z}-sequences. The resulting \textit{conformed} \textbf{z}-sequence then goes into our model for final output. We perform the edit on \textit{keyframes}, temporally sparse points in latent space, so that after the conform, we can interpolate between them for smooth, continuous output.

\paragraph{Process:} 
Below we outline an example process in more detail. This is merely a suggestion, as many different workflows could work, however this is the process we used to produce the example videos. 

In the following context  `\textit{render}' refers to i) saving the \textbf{z}-sequence to disk as a numpy array, ii) decoding the \textbf{z}-sequence with many snapshots of the model (from 28 different training iterations, spaced 1000 iterations apart) and saving out a video where the output of each snapshot is tiled into a grid (e.g. 7x4) and labelled with the corresponding training iteration. Rendering multiple snapshots in a grid on a single frame in this way gives us an overview of how the latent space has evolved across training iterations, and allows us to easily see and select the most aesthetically desirable snapshot(s) (we go deeper into the motivations behind this in the \hyperref[sec:snapshots]{\textit{Snapshots across time Appendix}}).

1) Take many (e.g. hundreds or thousands of) unbiased samples in latent space and render. This produces a video (and corresponding \textbf{z}-sequence) where each frame is an entirely different `random' image. This gives us an idea of what the model has learned, and how it is distributed. It also gives us an idea of how the distribution changes across subsequent training iterations, and which snapshots provide more aesthetically desirable images.
2) Edit the video in a NLE to remove undesirable (i.e.`bad') images or to bias the distribution (e.g. remove some frames containing flowers if there are too many, or duplicate frames containing bacteria if there's not enough of them etc.)
3) Run the script to conform the edit with the original \textbf{z}-sequence and re-render. This produces a new video (and corresponding \textbf{z}-sequence) where each frame is an entirely different `random' image, but which has hopefully a desired distribution (no `bad' images, and a desirable balance between different images).
4) Repeat steps 2-3 until we are happy with the distribution (one or two rounds is usually enough). Optionally apply varying amounts of noise in \textbf{z} to explore neighbourhoods of selected frames.
5) Load the final edited \textbf{z}-sequence (with desired distribution) and render many (e.g. tens or hundreds of) short \textit{journeys} interpolating between two or three random (or hand-picked) \textbf{z} (selected from the \textbf{z}-sequence). This produces tens or hundreds of short videos (and corresponding \textbf{z}-sequences) that contain smooth, slow interpolations between two or three \textit{keyframes} where the keyframes are chosen from our preferred distribution. This gives us an idea of how the model transitions between selected images. E.g. The shortest path from a mountain to a face might have to go through buildings, which might not be desirable, but inserting a flower in between might avoid the buildings and look nicer---both aesthetically and conceptually. 
6) Repeat step 5, honing in on journeys which seem promising, optionally applying varying amounts of noise in \textbf{z} to explore neighbourhoods of selected frames and journeys. The above steps produce an arsenal of video clips which can be further edited and joined in a NLE, and then conformed with the corresponding \textbf{z}-sequences to produce a final video. We have produced many hours worth of carefully constructed stories using this method.

\section{Conclusion}
Many aspects of this process can be improved; from theoretical, computational, and user experience points of view. We present this research as a first step in many, towards enabling users to meaningfully explore and control \textit{journeys} in high dimensional latent spaces to produce \textit{time-based media}, using and building upon industry standard tools and methods with which they may already be comfortable.

\subsubsection*{Acknowledgments}
This work has been supported by UK’s EPSRC Centre for Doctoral Training in Intelligent Games and Game Intelligence (IGGI; grant EP/L015846/1).

\bibliography{library}

\appendix

\clearpage
\section{Model architecture and data }
We applied the approach mentioned in this paper on a number of different models and architectures, however the primary test case we refer to (and from which we also show the results) is a GAN (specifically, a Progressively Grown GAN \cite{Karras2017}) trained on over 100,000 images scraped from the photo sharing website \textit{flickr}. The dataset is very diverse and includes images tagged with: art, cosmos, everything, faith, flower, god, landscape, life, love, micro, macro, bacteria, mountains, nature, nebula, galaxy, ritual, sky, underwater, marinelife, waves, ocean, worship and more. We include three thousand images from each category and train the network with \textit{no classification labels}. Given such a diverse dataset without any labels, the network is forced to try and organise its distribution based purely on aesthetics, without any semantic information. Thus in this high dimensional latent space we find directions allowing us to seamlessly morph from swarms of bacteria to clouds of nebula, oceanic waves to mountains, flowers to sunsets, blood cells to technical illustrations etc. Most interestingly, we can perform these transformations across categories \textit{while maintaining overall composition and form}.

\section{Video editing and conforming the edit}
\begin{figure}[h!]
	\includegraphics[width=\linewidth]{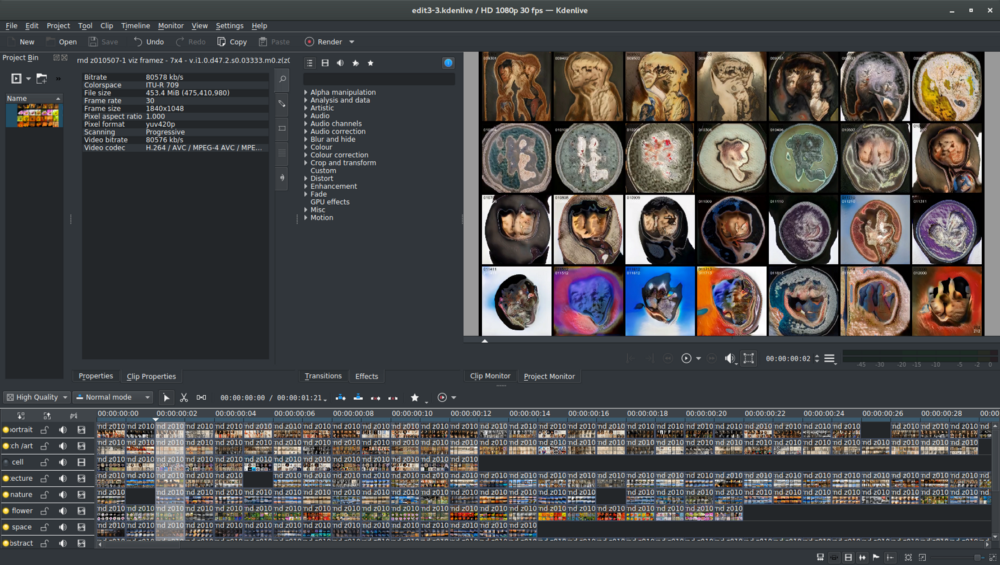}
	\caption{An example project in Kdenlive}
	\label{fig:kdenlive}
\end{figure}

We use the opensource Non Linear Video Editor \textit{Kdenlive} (Figure \ref{fig:kdenlive}) on Ubuntu. Unfortunately this editor lacks support for exporting the industry standard EDL. However, Kdenlive's native project file format is XML based. This allows us to write a Python based parser to load the project file, inspect the edits, retrieve the corresponding numpy \textbf{z}-sequences and conform by performing the same edits on them and exporting a new \textbf{z}-sequence. At this point the conform is very simple and only supports basic operations such as trimming, cutting and joining, and does not include cross-fades or other more advanced features or transitions. However, to implement such additional features is relatively trivial and left as future work (e.g. a cross-fade between two images in the NLE can be thought of as an interpolation between the two corresponding points in latent space).

\clearpage
\section{Interpolation}
\begin{figure}[h!]
	\includegraphics[width=\linewidth]{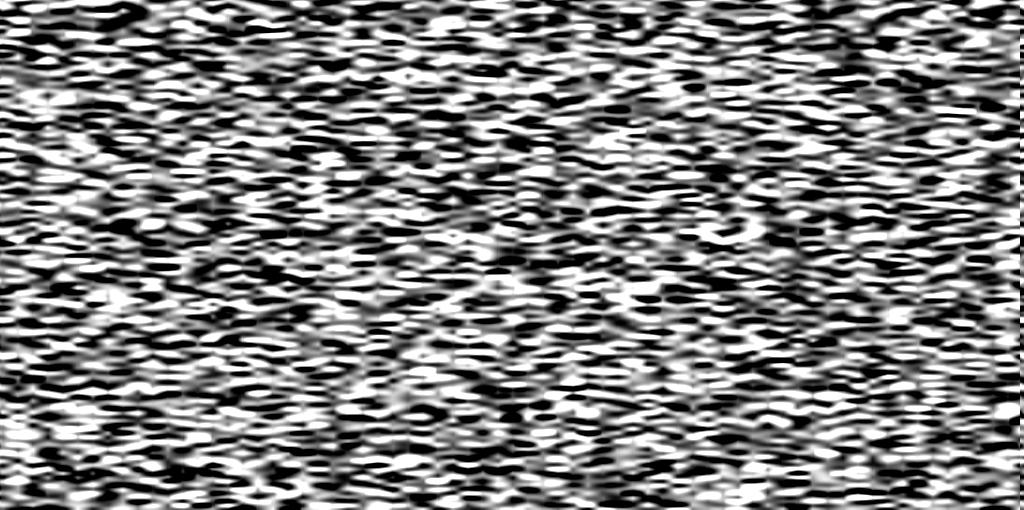}
	\caption{\textbf{z} sequence using \textit{spherical interpolation}}
	\label{fig:z_slerp}
\end{figure}

\begin{figure}[h!]
	\includegraphics[width=\linewidth]{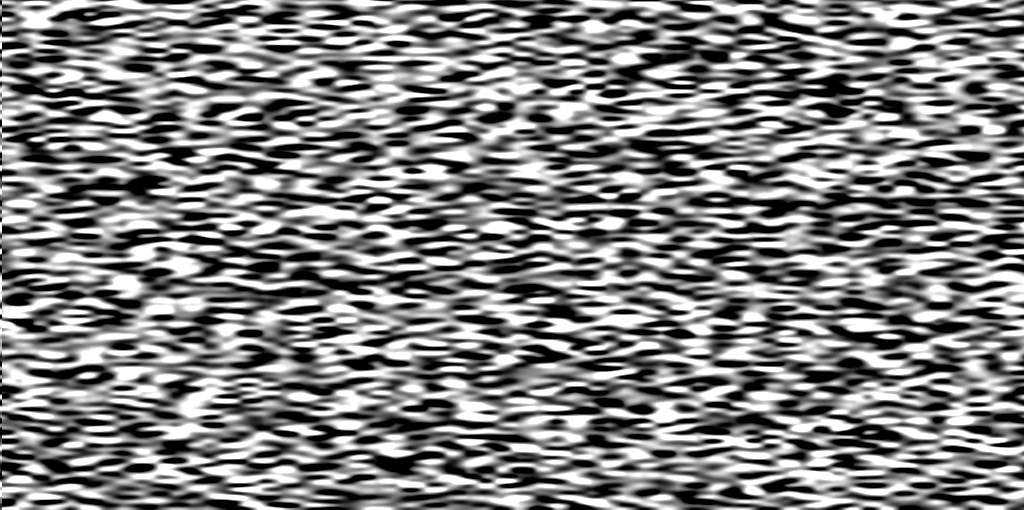}
	\caption{\textbf{z} sequence using \textit{physical interpolation}}
	\label{fig:z_physics}
\end{figure}

We use generative models with high dimensional (512D) multivariate Gaussian distributed latent spaces. Because these distributions are concentrated around the surface of a hypersphere \cite{Cook2011}, when we wish to interpolate between points in this space, we have to make sure that our trajectory stays within the distribution. A common solution is to use spherical instead of linear interpolation. However this produces visibly noticeable discontinuities in the movement of the output images due to sudden changes in speed and direction. The images below are two different \textbf{z} trajectories, i.e. journeys in latent space, created by interpolating between a number of arbitrary keyframes. In both images, a single pixel wide vertical slice represents a single \textbf{z} vector, and time flows left to right. Figure \ref{fig:z_slerp} visualises the results of spherical interpolation. We can see notch-like vertical artifacts that happen when the interpolation reaches its destination and we set a new target, creating a sudden change in speed and direction. To remedy this we introduce a simple physics based system, the results of which can be seen in Figure \ref{fig:z_physics}. In the high dimensional latent space we create a particle connected to both the surface of the hypersphere and the next destination point with damped springs. This ensures that the particle stays close to the distribution, but also moves without discontinuities at keyframes. 
    
\clearpage
\section{Snapshots across time}
\label{sec:snapshots}
\begin{figure}[ht!]
	\includegraphics[width=\linewidth]{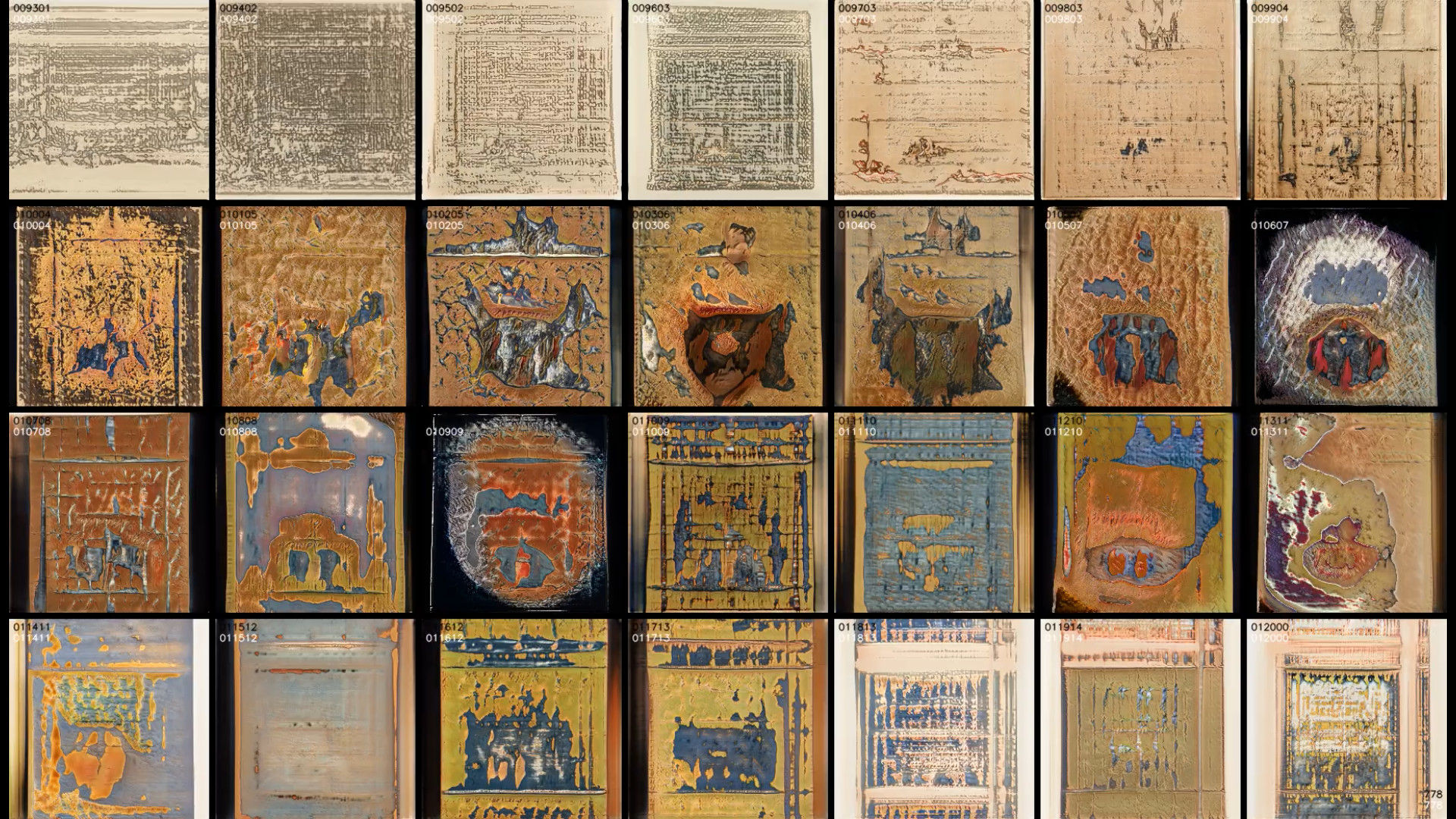}
	\caption{An example frame where the same \textbf{z}-vector is decoded from 28 snapshots spaced 1000 training iterations apart.}
	\label{fig:snapshot10}
\end{figure}

As the network trains, the latent space changes with each training iteration, to hopefully represent the data more efficiently and accurately. However a noticeable change across these iterations also includes transformations and \textit{shifts} in space. E.g. what may be an area in latent space dedicated to `mountains' at iteration 70K, might become `flowers' at iteration 80K, while `mountains' slide over to what used to be `clouds' (this is a bit of an exaggerated oversimplification). To investigate the effects of these transformations, we render the same \textbf{z}-sequence decoding from a number of different snapshots across subsequent training iterations (e.g. the last 28 snapshots spaced 1000 iterations apart), and we tile the outputs in a grid (e.g. 7x4) when saving a video. An example video can be seen at \url{https://www.youtube.com/watch?v=DVsf0ooqFWE} and Figures \ref{fig:snapshot1}-\ref{fig:snapshot10} show example frames.

\begin{figure}[ht!]
	\includegraphics[width=\linewidth]{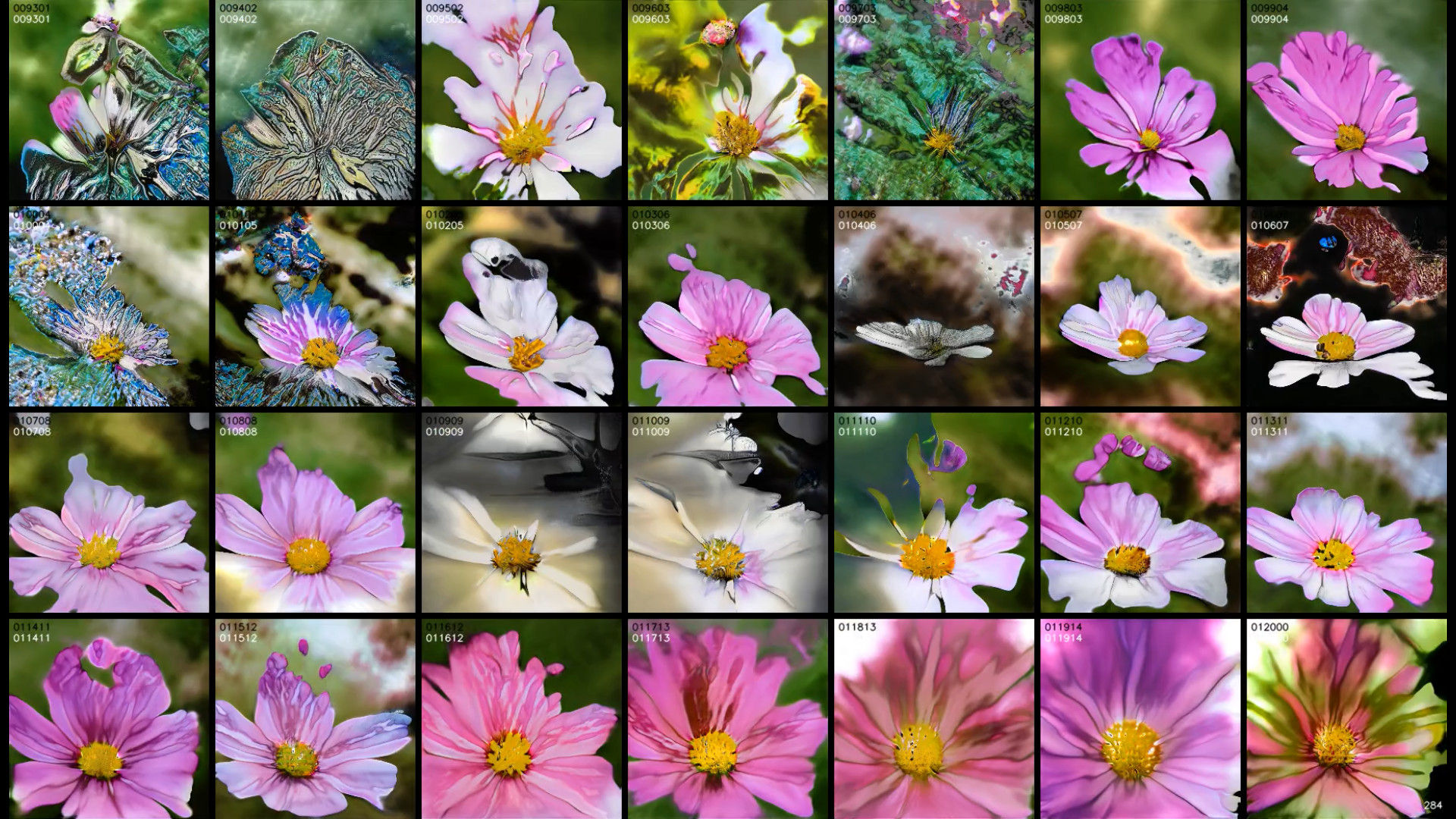}
	\caption{An example frame where the same \textbf{z}-vector is decoded from 28 snapshots spaced 1000 training iterations apart.}
	\label{fig:snapshot1}
\end{figure}

\begin{figure}[ht!]
	\includegraphics[width=\linewidth]{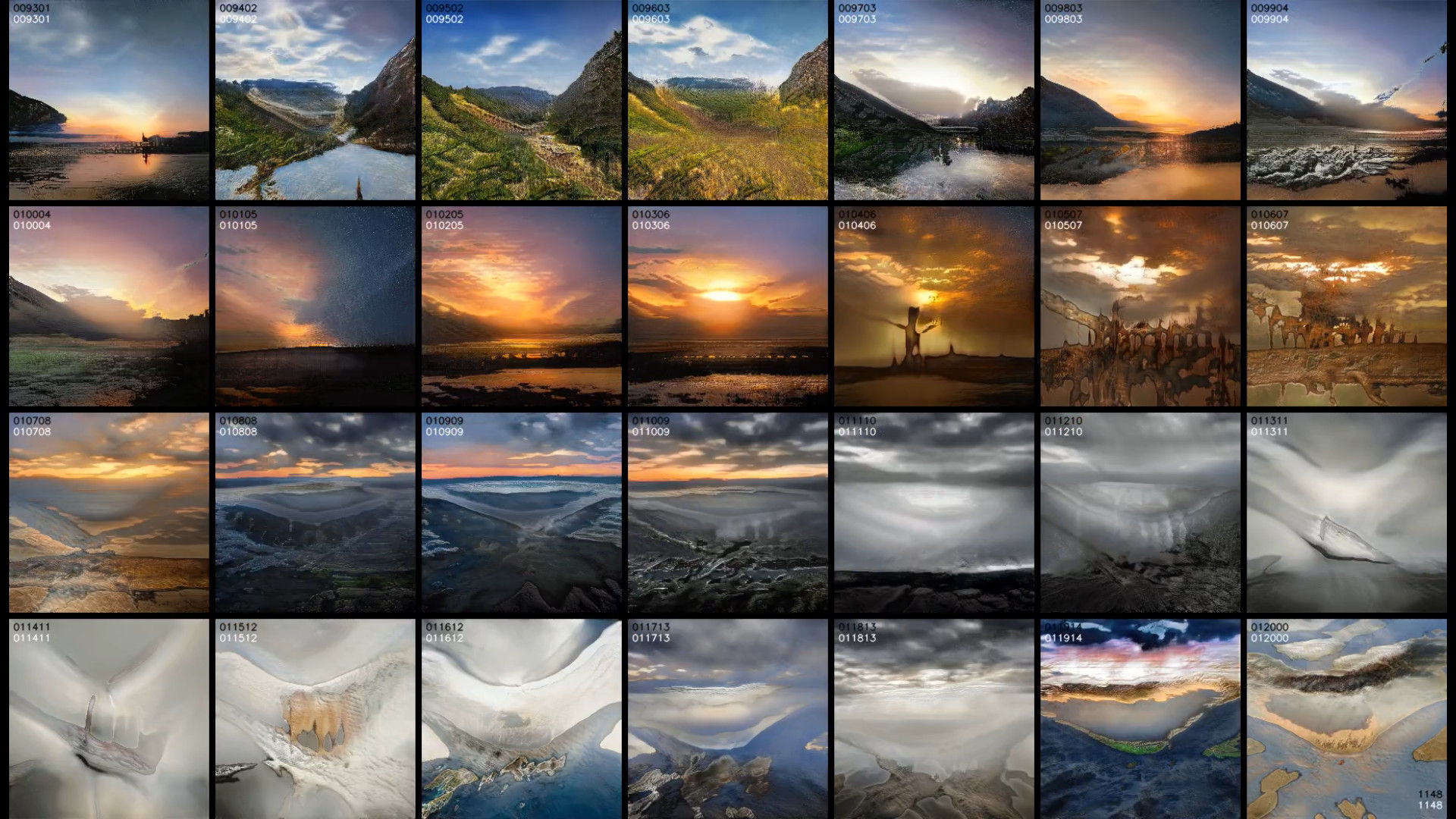}
	\caption{An example frame where the same \textbf{z}-vector is decoded from 28 snapshots spaced 1000 training iterations apart.}	
	\label{fig:snapshot2}
\end{figure}

Here, every tile within a frame is the \textit{same \textbf{z}-vector} decoded from a different \textit{snapshot in time (i.e. training iteration)}. The small number in the top left of each tile (in both black and white) is the iteration number. We can see in many cases the images are relatively similar with slight variations. In other cases there are more radical shifts, where earlier snapshots are hinting at generating one type of image while later snapshots are producing another for the same \textbf{z}-vector. Interestingly, even while semantically the images might be radically different, sometimes the overall form and composition is similar. E.g. in Figure \ref{fig:snapshot2} we can see that the space occupied by the current \textbf{z}-vector briefly gives way from mountains to flowers, however the images maintain the valley-like shape.  

When editing our videos in the NLE, we edit these videos containing the outputs from multiple tiled snapshots. This gives us an overview of the aesthetic qualities from the different training iterations, and allows us to choose the most aesthetically desirable snapshot(s) to use for our final output.

\begin{figure}[ht!]
	\includegraphics[width=\linewidth]{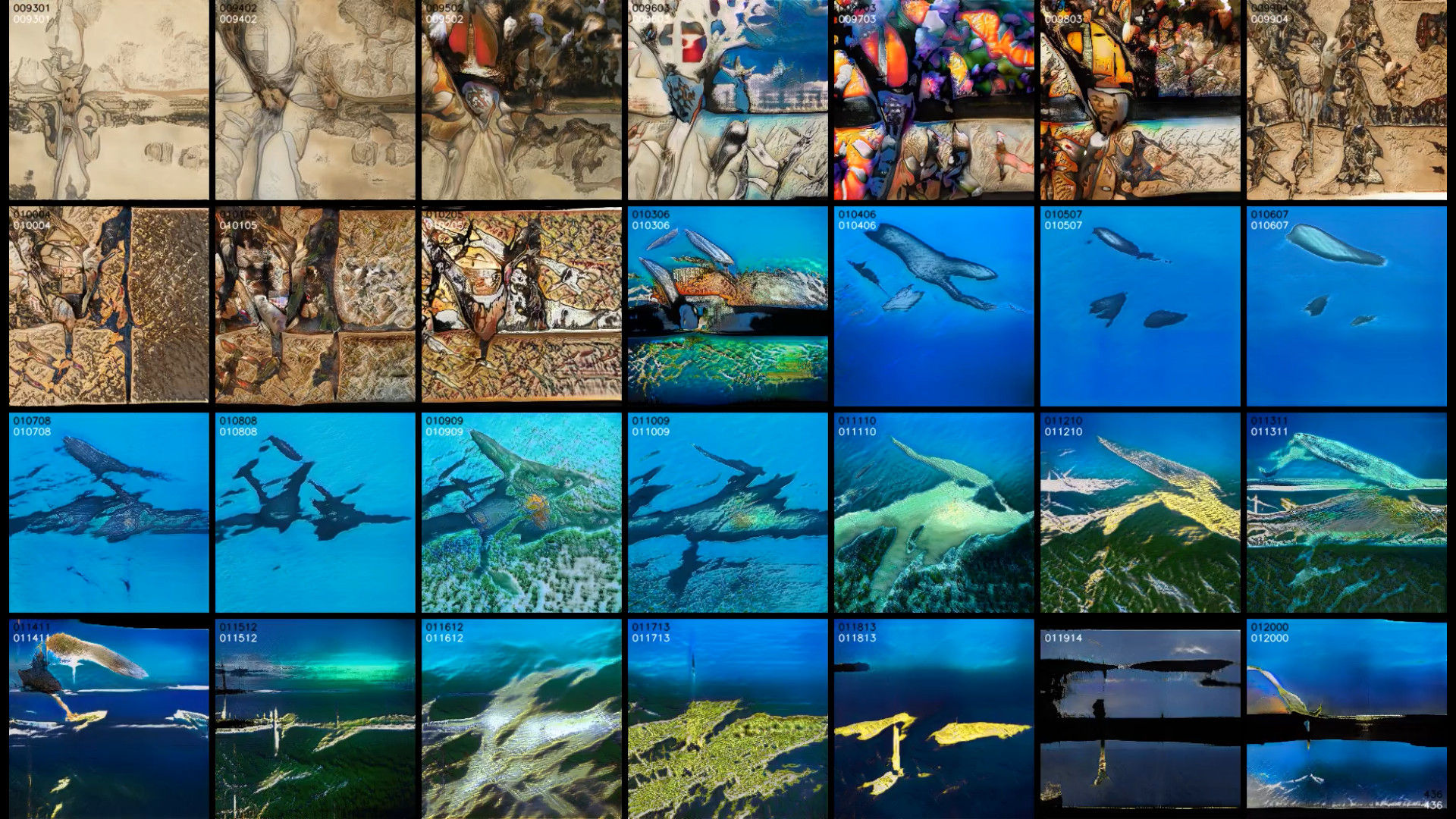}
	\caption{An example frame where the same \textbf{z}-vector is decoded from 28 snapshots spaced 1000 training iterations apart.}	
	\label{fig:snapshot1}
\end{figure}

\begin{figure}[ht!]
	\includegraphics[width=\linewidth]{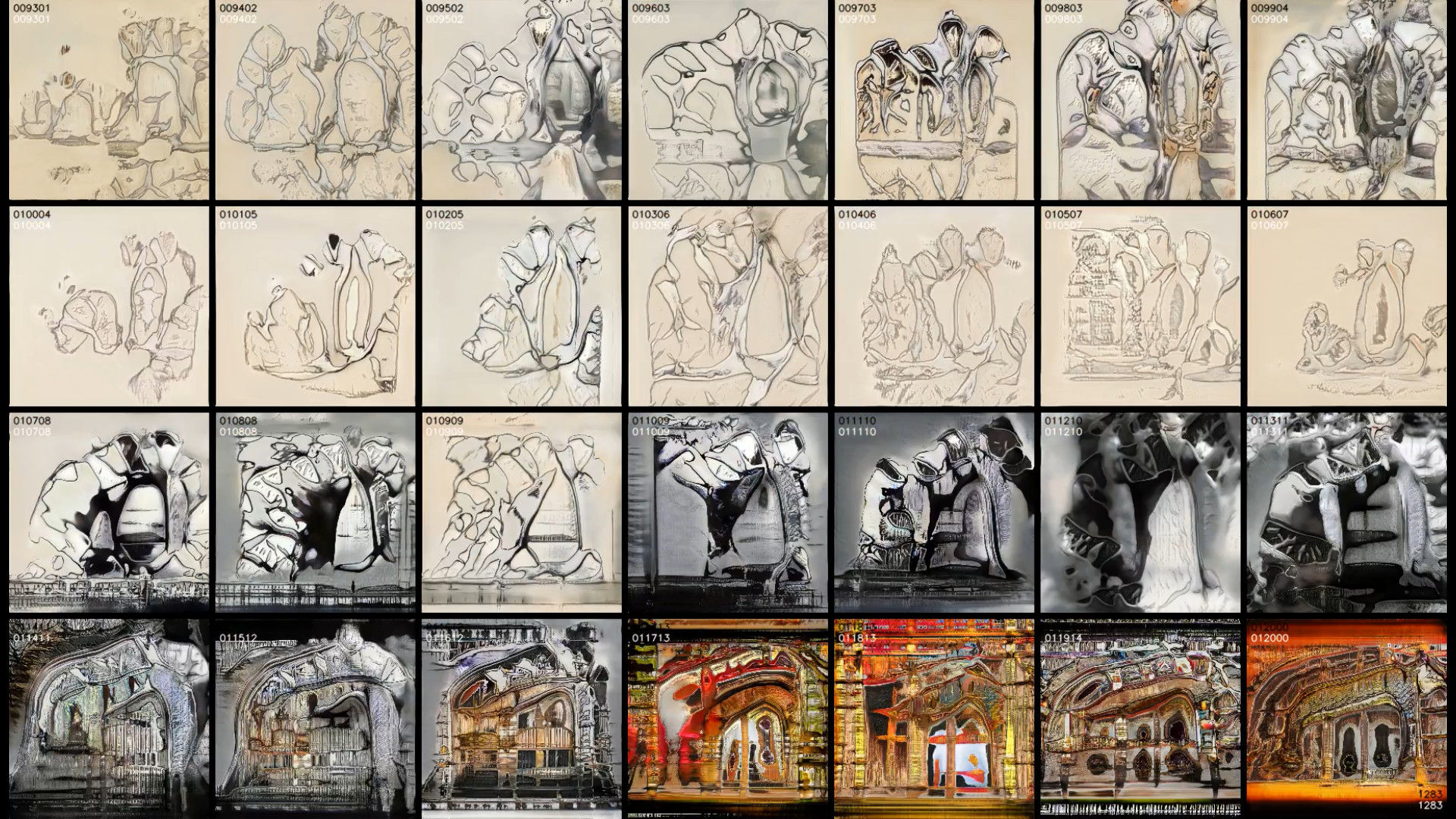}
	\caption{An example frame where the same \textbf{z}-vector is decoded from 28 snapshots spaced 1000 training iterations apart.}
	\label{fig:snapshot3}
\end{figure}

\begin{figure}[ht!]
	\includegraphics[width=\linewidth]{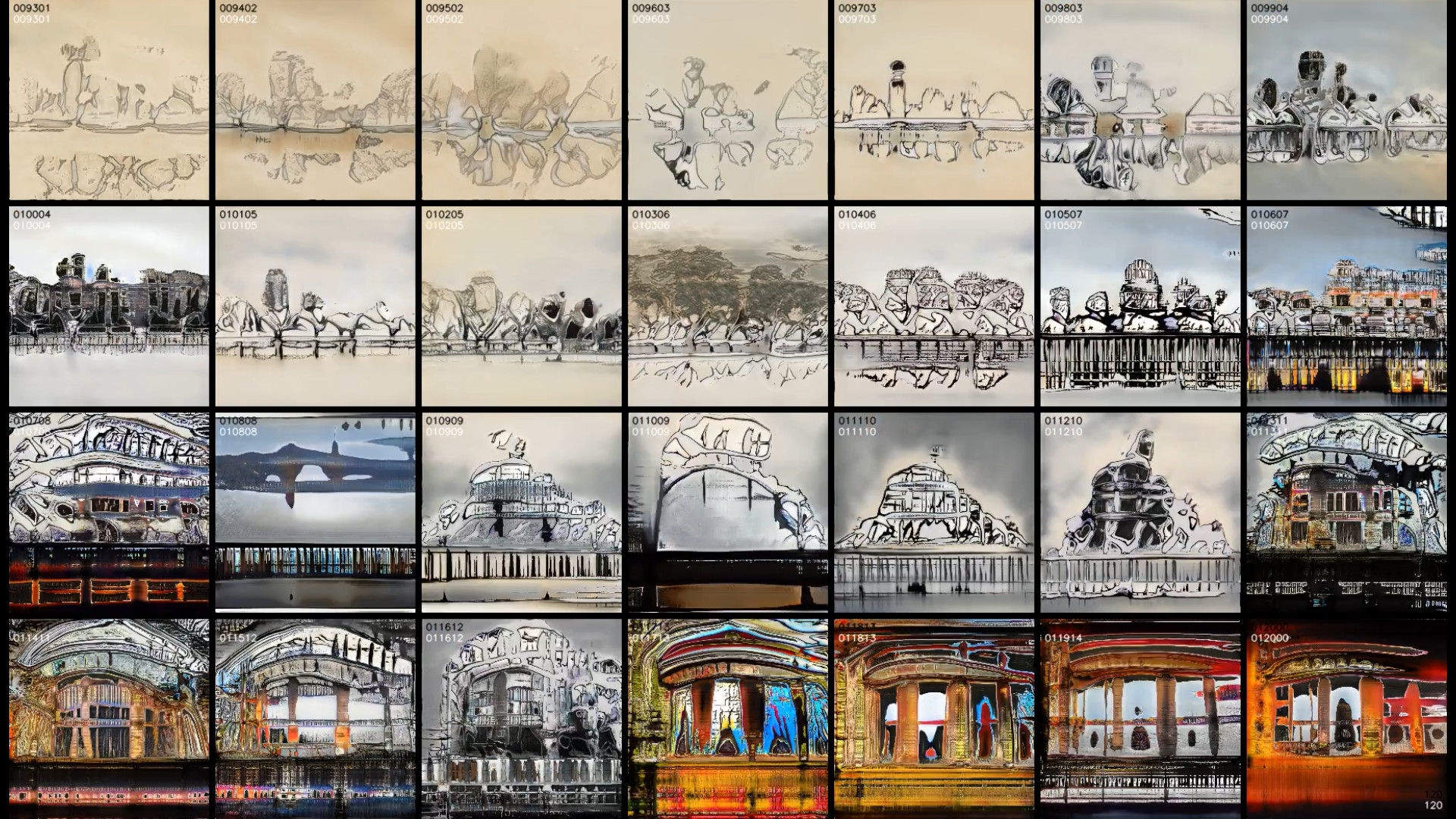}
	\caption{An example frame where the same \textbf{z}-vector is decoded from 28 snapshots spaced 1000 training iterations apart.}
	\label{fig:snapshot4}
\end{figure}

\begin{figure}[ht!]
	\includegraphics[width=\linewidth]{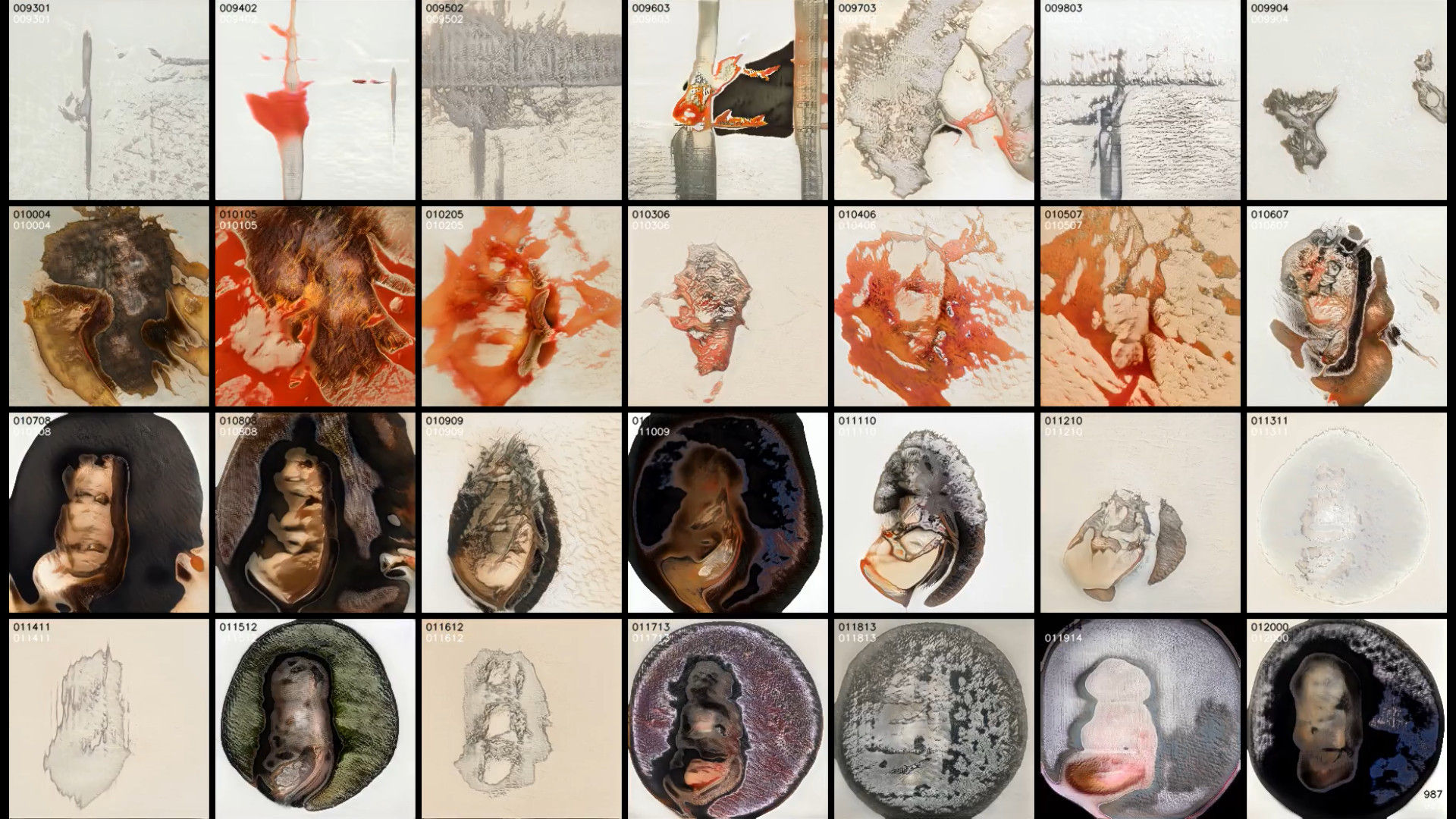}
	\caption{An example frame where the same \textbf{z}-vector is decoded from 28 snapshots spaced 1000 training iterations apart.}
	\label{fig:snapshot5}
\end{figure}

\begin{figure}[ht!]
	\includegraphics[width=\linewidth]{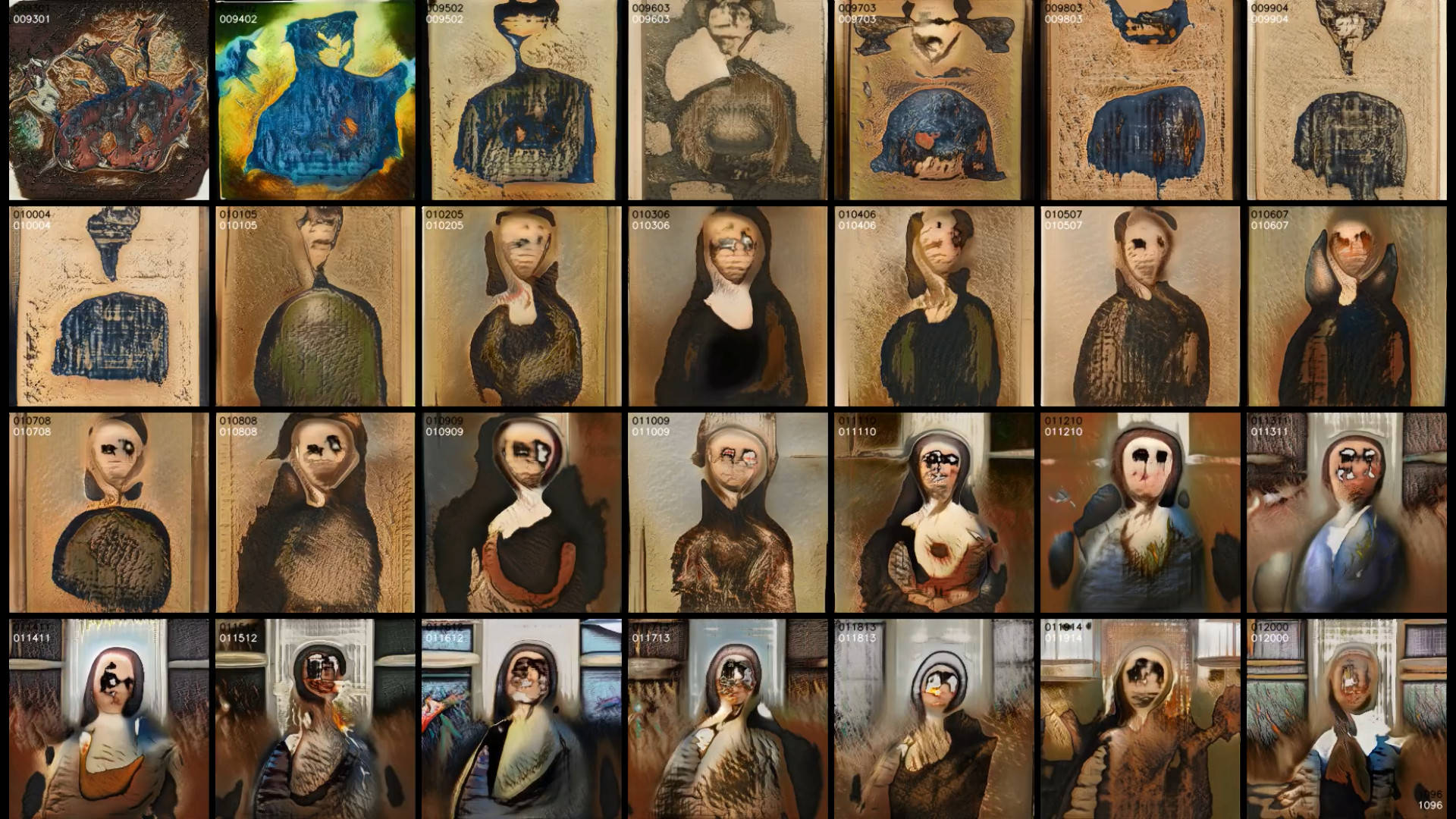}
	\caption{An example frame where the same \textbf{z}-vector is decoded from 28 snapshots spaced 1000 training iterations apart.}
	\label{fig:snapshot6}
\end{figure}

\begin{figure}[ht!]
	\includegraphics[width=\linewidth]{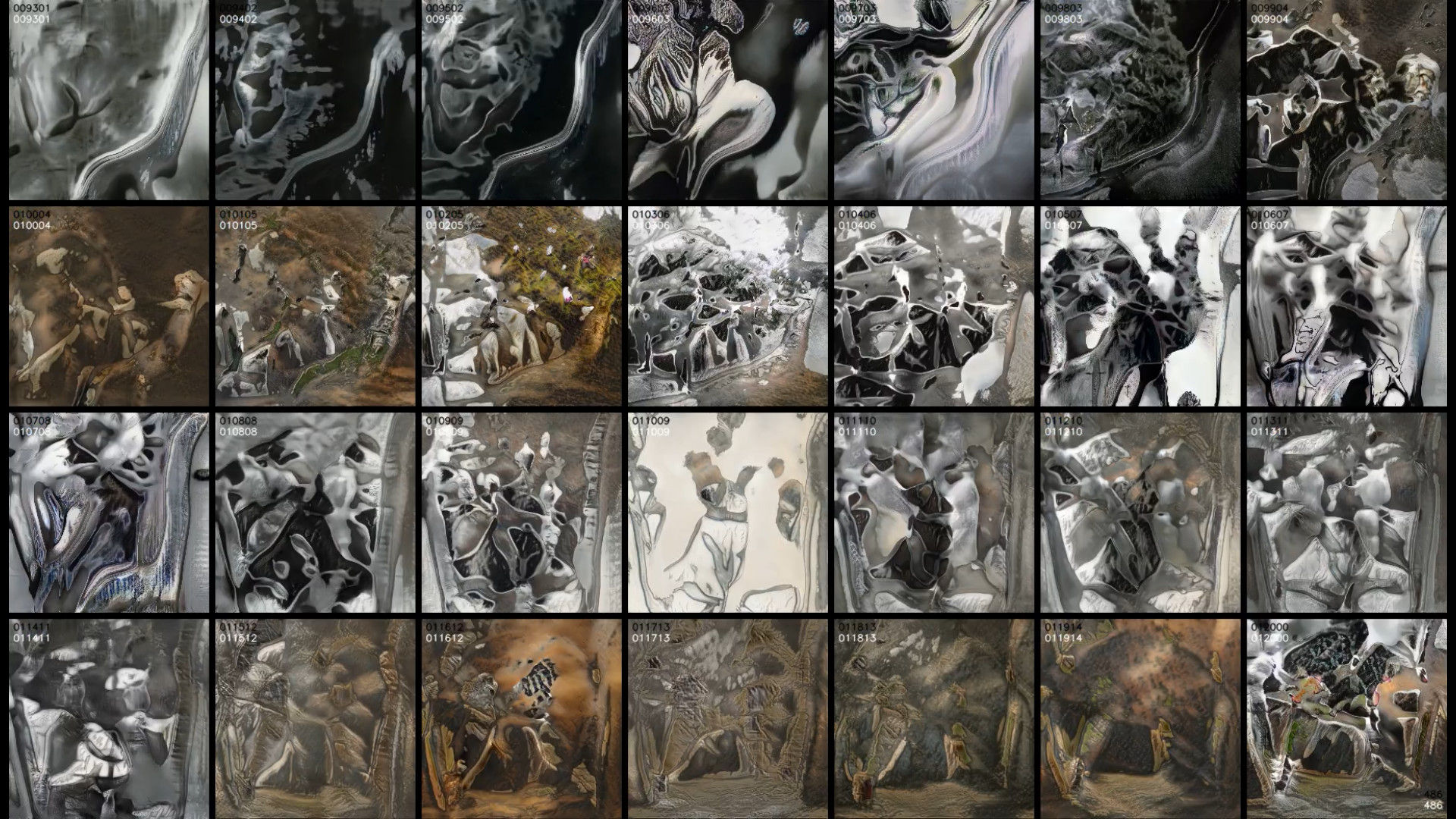}
	\caption{An example frame where the same \textbf{z}-vector is decoded from 28 snapshots spaced 1000 training iterations apart.}
	\label{fig:snapshot7}
\end{figure}

\begin{figure}[ht!]
	\includegraphics[width=\linewidth]{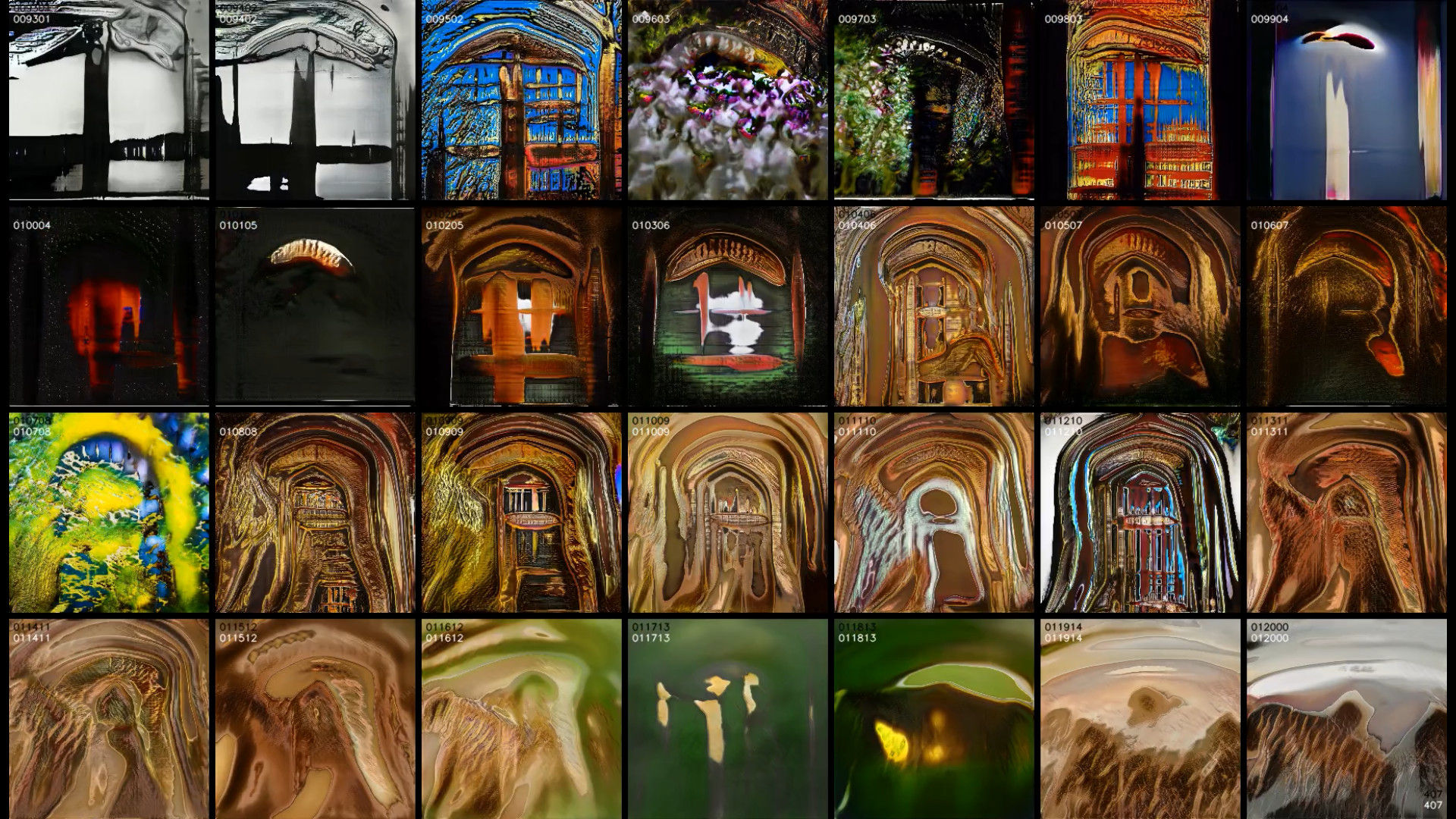}
	\caption{An example frame where the same \textbf{z}-vector is decoded from 28 snapshots spaced 1000 training iterations apart.}
	\label{fig:snapshot8}
\end{figure}

\begin{figure}[ht!]
	\includegraphics[width=\linewidth]{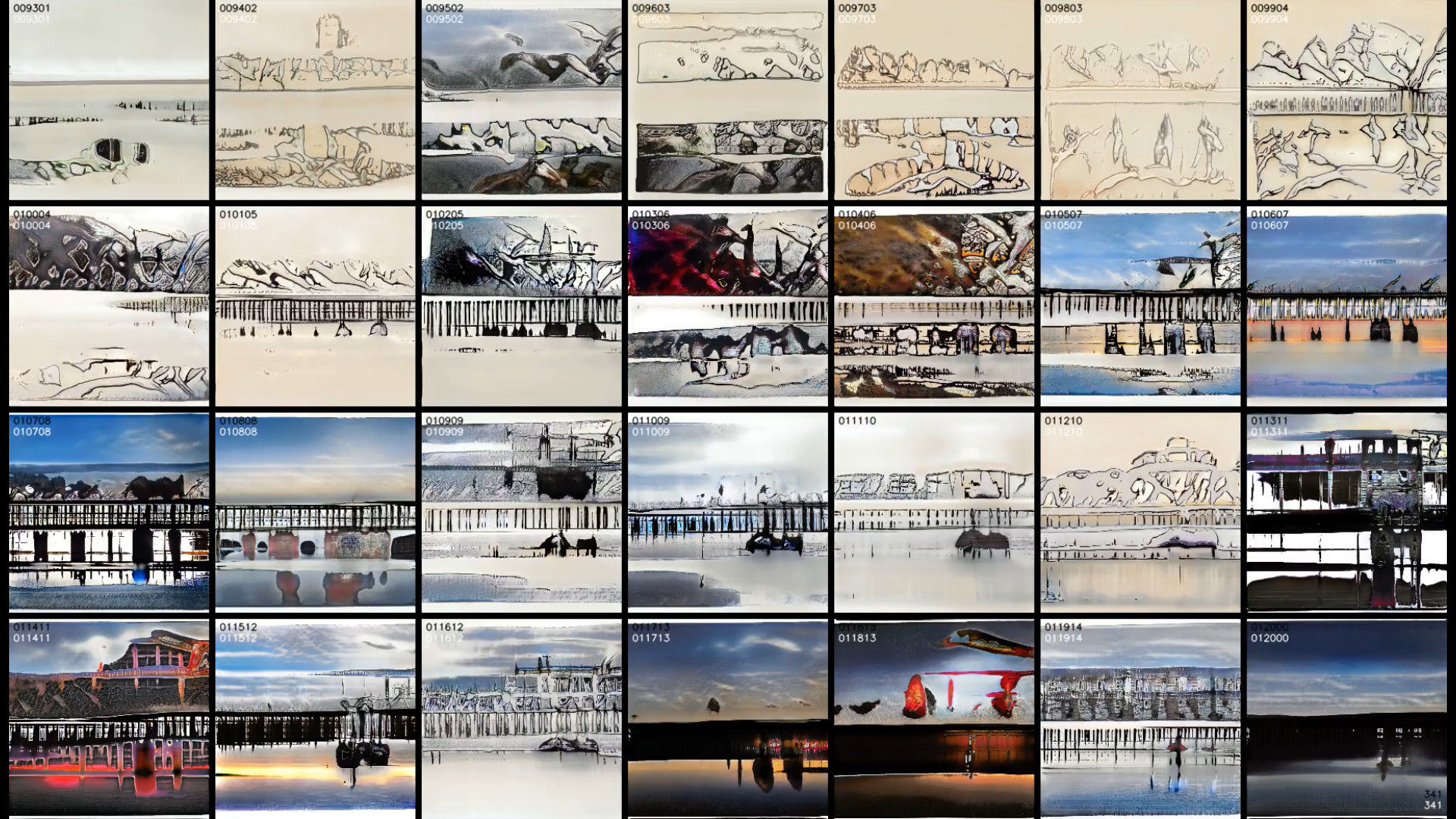}
	\caption{An example frame where the same \textbf{z}-vector is decoded from 28 snapshots spaced 1000 training iterations apart.}
	\label{fig:snapshot9}
\end{figure}

\end{document}